\documentclass{article}

\usepackage[final]{neurips_2019}
\immediate\write18{bibtex \jobname}
\usepackage{soul}

\usepackage[utf8]{inputenc} 
\usepackage[T1]{fontenc}    
\usepackage{hyperref}       
\usepackage{url}            
\usepackage{booktabs}       
\usepackage{amsfonts}       
\usepackage{nicefrac}       
\usepackage{microtype}      
\usepackage{times}
\usepackage{latexsym}
\usepackage{url}
\usepackage{booktabs}       
\usepackage{amsfonts}       
\usepackage{nicefrac}       
\usepackage{microtype}      
\usepackage{amssymb}
\usepackage{multirow}
\usepackage{multicol}
\usepackage{graphicx}
\usepackage{courier}
\usepackage{wrapfig}
\usepackage{adjustbox}
\usepackage{tabularx}
\usepackage{diagbox}
\usepackage{makecell}
\usepackage{caption}
\usepackage{enumitem}
\usepackage{titlesec}
\usepackage{natbib}
\usepackage{soul}
\usepackage{enumitem}
\usepackage[hang]{footmisc}

\usepackage{letltxmacro}

\title{Multi-domain Conversation Quality Evaluation \\ via \\ User Satisfaction Estimation}

\author{%
Praveen Kumar Bodigutla\\
 \texttt{pbodigut@amazon.com}\\
 \And 
Lazaros Polymenakos\\
 \texttt{polyml@amazon.com}\\
 \And
Spyros Matsoukas\\
 \texttt{matsouka@amazon.com}\\
}

\begin{document}

\maketitle

\begin{abstract}
An automated metric to evaluate dialogue quality is vital for optimizing data driven dialogue management. The common approach of relying on explicit user feedback during a conversation is intrusive and sparse. Current models to estimate user satisfaction use limited feature sets and employ annotation schemes with limited generalizability to conversations spanning multiple domains. To address these gaps, we created a new {\em Response Quality} annotation scheme, introduced five new domain-independent feature sets and experimented with six machine learning models to estimate {\em User Satisfaction} at both turn and dialogue level. 

Response Quality ratings achieved significantly high correlation ($0.76$) with explicit turn-level user ratings. Using the new feature sets we introduced, Gradient Boosting Regression model achieved best (rating [1-5])  prediction performance  on $26$ seen (linear correlation {\scriptsize $\sim$}$0.79$) and one new multi-turn domain (linear correlation $0.67$). We observed a $16\%$ relative improvement (68\% $\rightarrow$ 79\%) in binary (``satisfactory/dissatisfactory'') class prediction accuracy of a domain-independent dialogue-level satisfaction estimation model after including predicted turn-level satisfaction ratings as features.
\end{abstract}

\section{Introduction}

\label{sec:intro}

Automatic turn and dialogue level quality evaluation of end user interactions with Spoken Dialogue Systems (SDS) is vital for identifying problematic conversations and for optimizing dialogue policy using a data driven approach, such as reinforcement learning. One of the main obstacles to designing data-driven policies is the lack of an objective function to measure the success of a particular interaction.  Existing methods along with their limitations to measure dialogue success can be categorized into five groups: 1) Using sparse sentiment for end-to-end dialogue system training; 2) Using task success as dialogue evaluation criteria, which does not capture frustration caused in intermediate turns and assumes end user goal is known in advance; 3) Explicitly soliciting feedback from the user, which is intrusive and may cause dissatisfaction; 4) Using a popular approach, such as PARADISE \citep*{Walker:2000:TDG:973935.973945}, to predict dialogue-level satisfaction ratings provided by surveyed users, which has limited generalizability to diverse (novice and experienced) user population  \citep*{DBLP:journals/corr/abs-1905-04071}; and 5) Estimating per turn dialogue quality using trained Interaction Quality (IQ) \citep*{SCHMITT12.333} estimation models. Per turn discrete 1-5 scale IQ labels are provided by the annotator while keeping track of dialogue quality till that turn. 

Prior attempts to model user satisfaction at turn level as a continuous process evolving over time, employed either users \citep*{engelbrecht-etal-2009-modeling} or expert annotators \citep*{Higashinaka2010IssuesIP} to rate each turn in a dialogue. Both approaches suffered from manual feature extraction process, which IQ addressed. Features used in the IQ estimation models were derived from the current turn, the dialogue history, and output from three Spoken Language Understanding (SLU) components, namely: Automatic Speech Recognition (ASR), Natural Language Understanding (NLU), and the dialogue manager. IQ annotations provided by expert annotators are reliable in comparison to explicit ratings provided by the end users. Various models have been explored to predict IQ, including Hidden Markov Models \citep*{10.1007/978-1-4614-8280-2_27}, Support Vector Machines (SVM) \citep*{Schmitt2011ModelingAP}, Support Vector Ordinal Regression (SVOR) \citep*{6854195}, Recurrent Neural Networks (RNN) \citep*{Pragst2017}, and Long Short-Term Memory Networks (LSTM) \citep*{Rach2017InteractionQE}. 

Conversations users have with modern day SDS span multiple domains in a single dialogue session (Table \ref{multi-domain-example}). IQ annotation scheme which is developed using within domain conversations has limited generalizability to dialogues that span across multiple domains.

\begin{table*}[h!]
  \centering
  \resizebox{0.60\textheight}{!}{
  \begin{adjustbox}{max width=\textwidth}
  \bgroup
  \def\arraystretch{1.3}
  \begin{tabularx}{\linewidth}{XX}
    \toprule
    \multicolumn{1}{c}{\textbf{Utterance}}   & \multicolumn{1}{c}{\textbf{Comment}} \\
    \midrule
    \textbf{User:} Play latest hits. \textit{\{domain:\textbf{Music\}}}  &  \hspace{4mm} \textit{User makes a popular request}\\
    \textbf{System:} Shuffling from your playlist.  & \hspace{4mm} \textit{System addresses the request successfully} \\
    \textbf{User:}  Stop. Cancel my evening appointment. &\hspace{4mm} \textit{User barges-in} \\
                 \textit{\{domain:\textbf{Calendar}}\}  & \hspace{4mm} \\
    \textbf{System:} Sorry I don't know that one. &\hspace{4mm} \textit{System identifies un-actionable request}  \\
    \textbf{User:} Cancel my 7pm event if it is raining today. \textit{\{domain:\textbf{Calendar, Weather}\}} & \hspace{4mm} \textit{User paraphrases} \\
    \textbf{System:} Today in Seattle, there is 60\% chance of rain. & \hspace{1mm} {} \\ 
     Should I cancel your 7pm event - Park Visit? & \hspace{4mm} \textit{System Informs and asks a question.}\\
    \textbf{User:} Yes please. \textit{\{domain:\textbf{Global}\}} & \hspace{4mm} \textit{User confirms}\\
    \textbf{System:} Alright, 7pm event - Park visit Canceled.  & \hspace{4mm} \textit{System executes user request} \\
    \bottomrule
  \end{tabularx}
  \egroup
  \end{adjustbox}
  }
   \caption[Caption for LOF]{Conversation spanning multiple domains.\footnotemark}
      \vspace{-0mm}
  \label{multi-domain-example}
\end{table*}
\footnotetext{Due to confidentiality consideration, this example is not a real user conversation but authored to mimic the true dialogue.}

The IQ approach to rate turns in a dialogue incrementally, increases cognitive load on the annotators and lowers inter-annotator reliability. We also hypothesize that the features using in IQ estimation models are limited, and including additional contextual signals improves the performance of dialogue quality estimation models for both single-turn and multi-turn conversations\footnote{In single-turn conversations the entire context is expected to be present in the same turn, while in a multi-turn case the context is carried from previous turns to address user's request in the current turn.}. 

To address the aforementioned gaps in existing dialogue quality estimation methods, we propose an end-to-end User Satisfaction Estimation (USE) metric which predicts user satisfaction rating for each dialogue turn on a continuous 1-5 scale. To obtain consistent, simple and generalizable annotation scheme that easily scales to multi-domain conversations, we introduce turn-level Response Quality (RQ) annotation scheme.  To improve the performance of the satisfaction estimation models, we design five new feature sets, which are: 1) user request paraphrasing indicators, 2) cohesion between user request and system response, 3) diversity of topics discussed in a session, 4) un-actionable user request indicators, and 5) aggregate popularity of domains and topics across the entire population of the users. 

Using RQ annotation scheme, annotators rated dialogue turns from $26$ single-turn and multi-turn sampled Alexa domains\footnote{A multi-turn domain supports multi-turn conversations whereas a single-turn domain treats each user query as an independent request where context is not carried forward from previous turns.} (e.g., \textit{Music, Calendar, Weather, Movie booking}). We trained USE machine learning models using the annotated RQ ratings. To explain model predicted rating using features' values, we experimented with four interpretable models that rank features by their importance. We benchmarked performance of these models against two state-of-the-art dialogue quality prediction models. Using ablation studies, we showed improvement in the best performing USE model's performance using new contextual features we introduced.

In order to test the generalization performance of the model, we reserved one ``new'' multi-turn Alexa domain in the test set. Furthermore, we trained a dialogue-level satisfaction estimation model with explicit session-level ratings provided by both novice and experienced users on multi-turn dialogues from $14$ domains. We evaluated the impact of including turn-level satisfaction ratings as features on the performance of this domain-independent dialogue-level satisfaction estimation model.  

The outline of the paper is as follows: Section 2 introduces the Response Quality annotation scheme and discusses its effectiveness in terms of predicting turn-level user satisfaction ratings. Section 3 summarizes the Response Quality annotated data, discusses dialogue-level satisfaction estimation using ratings provided by novice \& experienced users and presents our experimentation setup. Section 4 provides an empirical study of turn and dialogue-level satisfaction estimation models' performance and discusses results from feature ablation study. Section 5 concludes.

\section{Response Quality Annotation}
We designed the Response Quality (RQ) annotation scheme to generate training data for our turn-level User Satisfaction Estimation (USE) model. 
\label{sec:annotation}
\subsection{RQ annotation Scheme and Comparison with IQ}
In RQ, similar to IQ annotations, annotators listened to raw audio and provided per turn's system RQ rating on a 5-point scale.  The scale we asked annotators to follow was: $1$=Terrible (system fails to understand and fulfill user's request), $2$=Bad (understands the request but fails to satisfy it in any way), $3$=OK (understands users request and either partially satisfies the request or provides information on how the request can be fulfilled), $4$=Good (understands and satisfies the user request, but provides more information than what the user requested or takes extra turns before meeting the request), and $5$=Excellent (understands and satisfies user request completely and efficiently).  Using a $5$ point scale  over a binary scale provides an option to the annotators to factor in their subjective interpretation of the extent of success or failure of system's response to satisfy a user's request.

Annotators rated conversations that spanned multiple domains. They were instructed to use the follow-up feedback from the user (e.g., user expresses frustration or rephrases an initial request) in making judgements. Unlike IQ annotation scheme, we removed the constraint on the annotators to keep track of the quality of dialogue so far while determining RQ ratings for a given turn. This relaxation in constraint, coupled with making full conversation context available to the annotators, reduced the cognitive load on them. This simplified annotation scheme not only helped in scaling RQ to multiple domains but also enabled precise identification of defective turns which is not straightforward in the case of IQ where an individual turn's IQ rating depends on the prior turns' ratings.

\subsection{Inter Annotator Agreement (IAA) and Correlation with user satisfaction rating \label{sec:IAA}}
We conducted a user study to verify the accuracy of RQ and IQ \citep*{SCHMITT12.333} annotation process. In the study, eight users were asked to achieve $30$ pre-determined goals. Dialogue sessions to achieve the given goals spanned over six single-turn and two multi-turn Alexa domains. For $15$ out of the $30$ goals, we asked the users to provide satisfaction rating on a discrete (1-5) scale based on turn's system response. For the remaining $15$ goals, we asked the users to rate each turn incrementally based on their perception of interaction so far. Then we sent the same utterances to six annotators (3 annotators per annotation type) for obtaining RQ (950 turns) and IQ (700 turns) annotations. The IQ annotations are less in number in comparison to RQ because we excluded turns where the users were unsure about the per-turn incremental satisfaction ratings they provided. We did not impose any restriction on the number of turns required to achieve a given goal, which led to an unequal number of turns per dialogue session across users who were assigned the same goal.

Since IQ annotation guidelines suggest rating unrelated queries independently, the annotators providing IQ annotations, were also instructed to identify independent interactions within a given dialogue session. Specifically, annotators marked the beginning of a new interaction if they felt that the user's goal in the current turn is unrelated to the one he/she tried to achieve in the previous turn.

We found that the $[1-5]$ scale RQ ratings provided by $3$ annotators were highly correlated (Spearman's rho $0.94$)\footnote{We also asked 3 novice annotators to provide RQ ratings for 400 single and multi-turn conversations' turns, the IAA was still high with mean Spearman's rho of 0.75.} with each other, suggesting high IAA. The mean RQ ratings were significantly (at 95\% confidence interval) correlated ($0.76$) with surveyed user satisfaction ratings with system's response. In the case of IQ ratings, mean IAA and correlation with user ratings dropped to $0.20$  and $0.32$ respectively, suggesting limited generalizability of IQ annotation scheme to multi-domain conversations. 

We used cohen's kappa to measure IAA between IQ annotators on the binary annotations they provided to indicate the beginning of an interaction within a dialogue session. Low IAA (cohen's kappa $0.16$) achieved on these binary annotations, indicate ambiguity faced by the annotators in-terms of identifying end user's goal from a sequence of turns within a dialogue.

\section{Data and Experimental Setup} 
This section describes our turn-level and dialogue-level dataset, details the list of features derived from various signals, and explains our experimentation setup.
\subsection{Turn-level Response Quality Data}
\begin{figure}[!h]
  \begin{center}
    \includegraphics[width=0.75\textwidth]{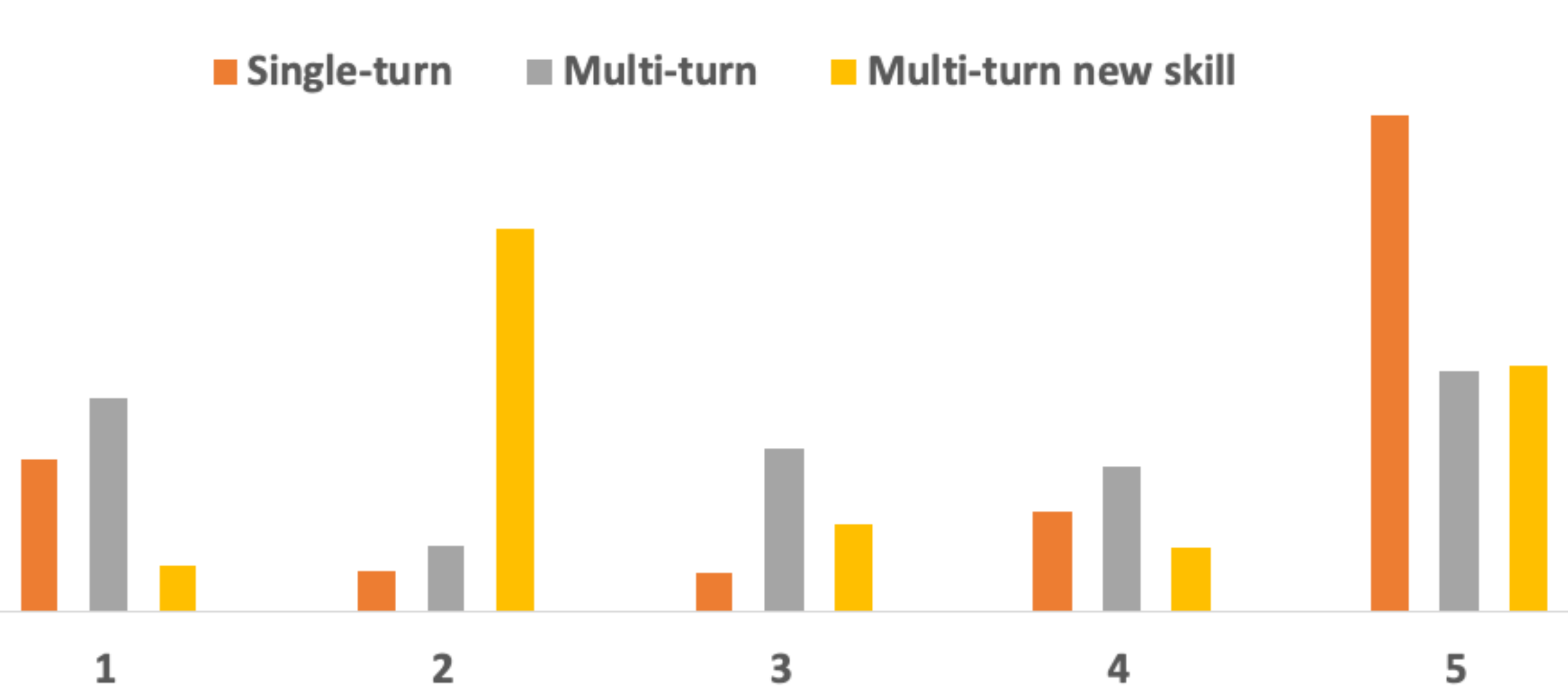}
  \end{center}
  \caption{Distribution of RQ annotation for single-turn, multi-turn domains and new multi-turn application. Exact percentage on y-axis is masked for confidentiality}
  \label{fig:rating-distribution}
\vspace{-0.5cm}
\end{figure}

\vspace{0.0cm}
\label{subsec:data}
To demonstrate that our RQ annotation scheme and predictive models are domain-independent and effective for both single-turn and multi-turn dialogues, we used 30,500 dialogue turns randomly sampled from $26$ single-turn ($90\%$) and multi-turn ($10\%$) domains that are representative of end user interactions with Alexa. The imbalance towards single-turn dialogues is due to annotation priority. We also tested model's generalization performance on $200$ dialogue turns sampled from a ``new'' multi-turn goal oriented application. Figure \ref{fig:rating-distribution} shows the diversity in rating distribution between the single-turn and multi-turn dialogues.

\label{subsec:feat}

\subsection{Features}
\begin{figure}[!h]
  \begin{center}
    \includegraphics[width=0.75\textwidth]{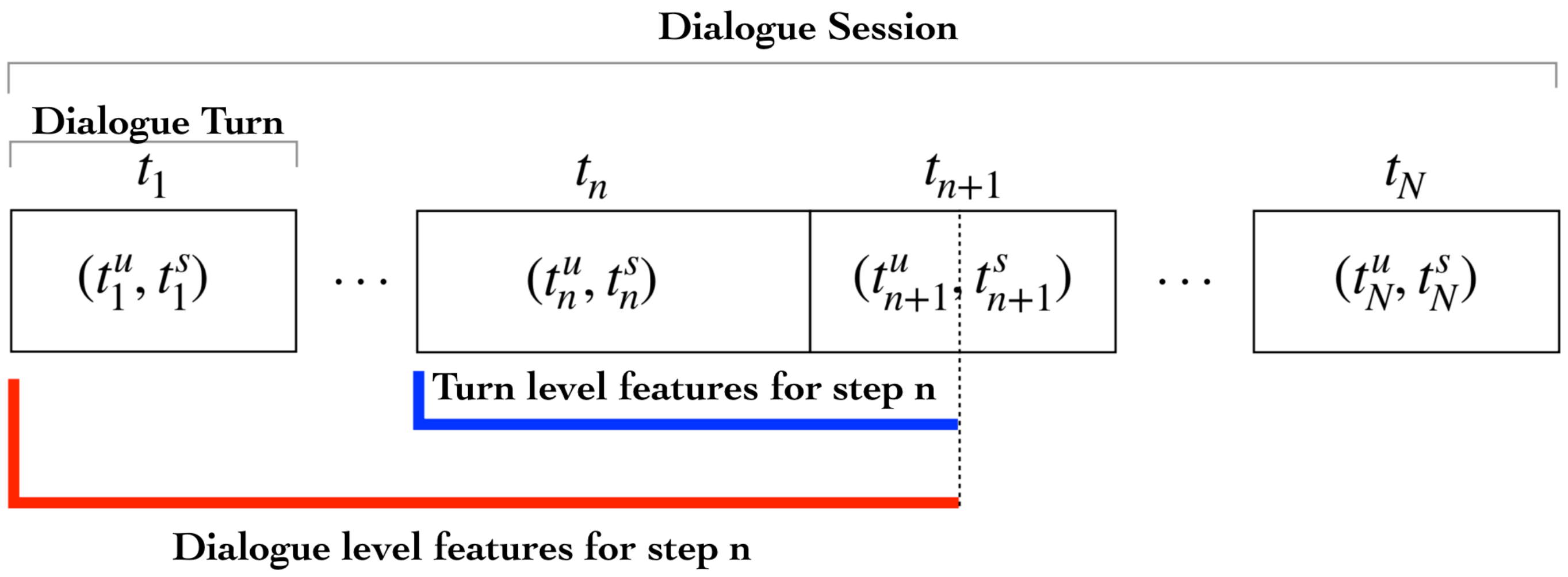}
  \end{center}
  \caption{dialogue and turn definitions. The blue and red lines indicate the history used for generating turn-based and dialogue-based features for the user satisfaction estimation on turn $t_n$.}
  \label{fig:turn-dialog-parameters}
\vspace{-0.5cm}
\end{figure}

To estimate the turn level user satisfaction score, we used features derived from turn, dialogue context, and Spoken Language Understanding (SLU) components' output similar to turn level IQ prediction models \citep*{Ultes2017DomainIndependentUS}. As shown in Figure \ref{fig:turn-dialog-parameters}, we define a dialogue turn at time $n$ as $t_n=(t_n^u, t_n^s)$, where $t_n^u$ and $t_n^s$ represent the user request and system response on turn $n$. A dialogue session of $N$ turns is defined as ($t_1$:$t_N$).  To improve the performance of the USE models across single and multi-turn conversations spanning multiple domains, we introduced the following $5$ sets of domain-independent features: 

\begin{enumerate}[nolistsep]
\item \textbf{User request paraphrasing} -- Calculated by measuring syntactic and semantic (NLU predicted intent) similarity between consecutive turns' user utterances. 

\item \textbf{Cohesion between response and request} -- Cohesiveness of system response with a user request is computed by calculating jaccard similarity score between user request and system response. System response \textit{``Here is a sci-fi movie''} will get higher cohesion score over \textit{``Here is a comedy movie''}, if the user request was \textit{``recommend a sci-fi movie''}.

\item \textbf{Aggregate topic popularity} -- Usage statistics such as aggregate domain and intent usage count and ratio of usage count to number of customers, provides us a prior on the popularity of a topic across all users of Alexa.

\item \textbf{Un-actionable user request} -- Identifies if the user request could not be fulfilled, by searching for phrases indicating an apology and negation in system response (e.g., \textit{``sorry I don't know how to do that''}). 

 \item \textbf{Diversity of topics in a session} --  This dialogue level feature is calculated using the percentage of unique intents till the current dialogue turn.
\end{enumerate}
\label{subsec:feat}

\subsection{Dialogue-level satisfaction estimation using predicted turn-level ratings}
To show that predicted turn-level satisfaction ratings are effective in estimating overall dialogue level satisfaction we trained dialogue-level quality estimation models to predict explicit dialogue-level ratings provided by users. Dialogue-level labels were obtained from ratings provided by $10$ users ($5$ novice and $5$ experienced)  who were asked to achieve $40$ multi-turn goals spanning $14$ \{commercial conversation agent's\} domains (example goals in Appendix Table \ref{user-study-goals}). 

Since earlier attempts to estimate explicit dialogue-level satisfaction ratings did not generalize to different user population (see section \ref{sec:intro}), we collected data from users belonging to both ``novice'' and ``experienced'' groups. A novice user has minimal experience conversing with the SDS and he/she has never used the functionality provided by the $14$ domains prior to the study. An experienced user is a seasoned user of Alexa and its domains. 

\begin{figure}[!h]
  \begin{center}
    \includegraphics[width=0.60\textwidth, height=5cm]{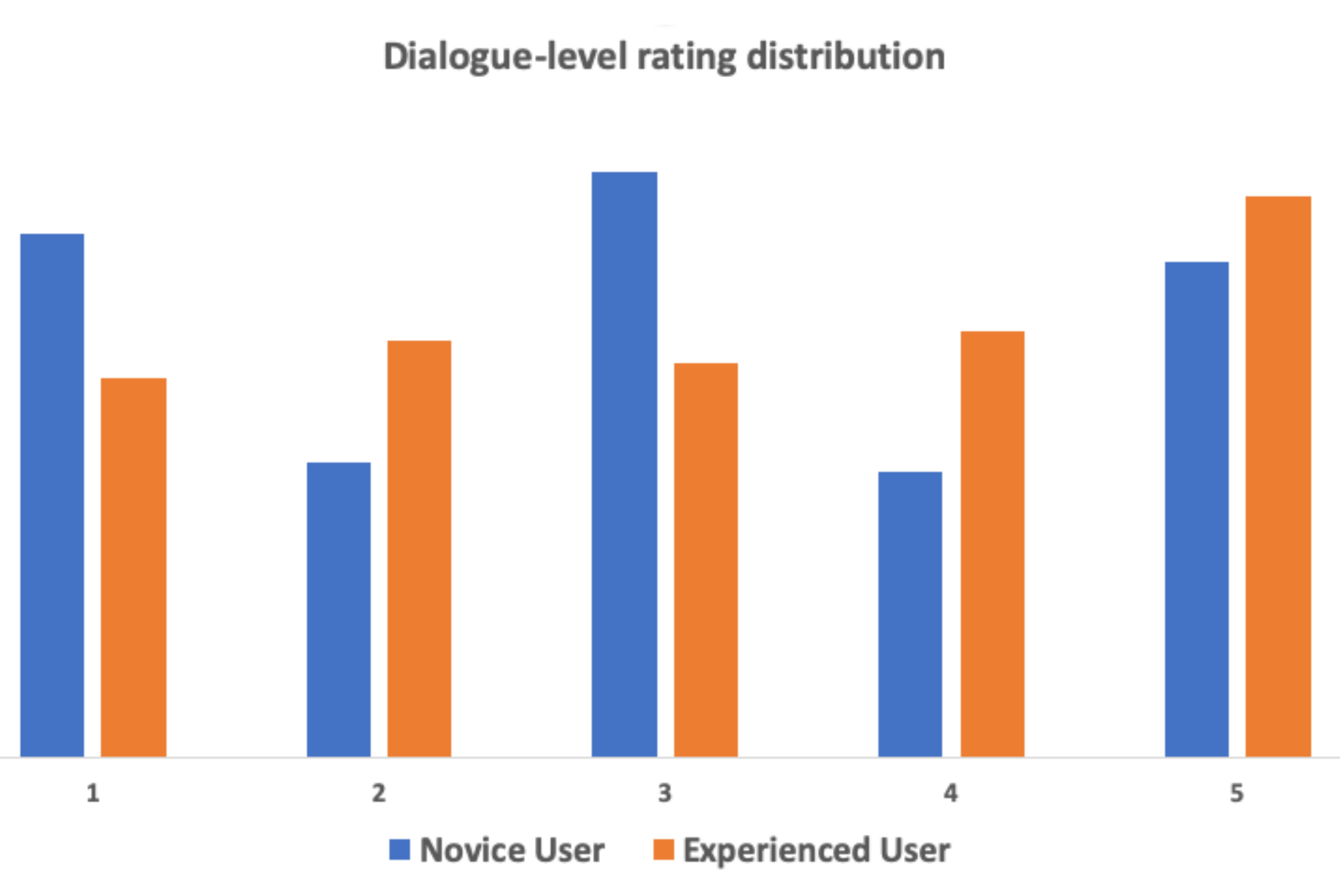}
  \end{center}
  \vspace{-1mm}
  \caption{Distribution of dialogue level satisfaction ratings. Exact percentage on y-axis is masked for confidentiality}
  \label{fig:dialog-rating-distribution}
\vspace{-0.5cm}
\end{figure}

Users provided their satisfaction rating with the dialogue on a discrete $[1-5]$ scale at the end of each session, irrespective of the outcome. The rating scale we asked the annotators to follow was $1$=Very dissatisfied, $2$=Dissatisfied, $3$=Moderately Satisfied (or Slightly dissatisfied), $4$=Satisfied and $5$=Extremely Satisfied. We collected $1,042$ dialogue-level satisfaction ratings in total.  $45\%$ of these ratings came from novice users and the rest were provided by the experienced users. Figure \ref{fig:dialog-rating-distribution} shows the distribution of dialogue level ratings from both sets of users.

Features to train the dialogue level satisfaction model included turn-level features computed on the last turn ($t_N$), aggregate statistics (e.g., average ASR confidence score,  number of barge-ins) computed over all turns ($t_1$:$t_N$) in a dialogue session and average estimated turn-level satisfaction rating calculated across all turns in the same session. 

\label{subsec:feat}

\subsection{Experimental Setup}
This section describes the experimental setup we used for training and evaluating turn and dialogue level satisfaction estimation models.
\label{subsec:setup}
\subsubsection{Turn-level satisfaction estimation}
We obtained a turn's RQ rating by averaging the discrete 1-5 labels provided by $3$ annotators. Hence, we considered regression models for experimentation which predicted satisfaction rating on a continuous 1-5 scale.  To achieve interpretability, in our experiments we selected four models - LASSO \citep*{10.2307/2346178}, Decision Tree Regression \citep*{Agrawal:1993:MAR:170036.170072},  Random Forest Regression \citep*{breiman2001random} and Gradient Boosting Regression \citep*{friedman2001greedy} that rank features by their importance. For benchmarking we used Multi-layer Perceptron (MLP) \citep*{gardner1998artificial} and Support Vector Regression (SVR) \footnote{Recurrent Neural Networks and Long Short-term Memory Networks \citep*{Rach2017InteractionQE} \citep*{NIPS1996_1238} showed non-significant improvement ($\leq0.02$ difference in Spearman's rho) over state-of-the-practice MLP and SVM models in predicting IQ.}. Data was randomly split into training ($60\%$), validation ($20\%$) and test ($20\%$) sets, however turns from the ``new'' multi-turn application were included only in the test set.

\subsubsection{Dialogue-level satisfaction estimation}
We trained the same four interpretable regression models (LASSO, Decision Tree, Random Forest  and Gradient Boosting) to predict explicit dialogue level ratings on a continuous $[1-5]$ scale. Since ratings obtained at dialogue level from users are sparse ($3.5\%$ of total number of turn-level RQ annotations), we performed 9-fold cross validation on randomly sampled $90\%$ of data and we evaluated the performance of the trained model on remaining $10\%$ test data.

\subsubsection{Evaluation Criteria}
We used Pearson's linear correlation coefficient ($r$) for evaluating each model's 1-5 prediction performance.  For the use case to identify problematic turns from an end user's perspective, it is sufficient to identify {\em satisfactory} (rating $\geq$3) and {\em dissatisfactory} (rating $<3$) interactions. For evaluating turn-level satisfaction models, we used F-score for the dissatisfactory class as the binary classification metric. Identifying dissatisfactory turns is of more importance and is in general a difficult task as majority of turns belong to the satisfactory class (Figure \ref{fig:rating-distribution}). We used {\em satisfactory} and {\em dissatisfactory} binary class prediction accuracy as an additional metric to evaluate the performance of dialogue-level satisfaction estimation models, since unlike turn-level data, the dialogue ratings are not concentrated around one class (Figure \ref{fig:dialog-rating-distribution}).

\section{Results and Analysis}

In this section we present a performance analysis of both turn-level and dialogue-level satisfaction estimation models.
\subsection{Turn-level satisfaction estimation results}
\label{sec:turn-level-satisfaction-results}
\begin{table*}[h!]
\captionsetup{font=small}
\centering
\resizebox{0.60\textheight}{!}{
  \begin{adjustbox}{max width=\textwidth}
  \bgroup
  \def\arraystretch{1.1}
\begin{tabular}{|c|c|c|c|c|c|c|}
\hline
\multicolumn {1}{|c}{} & \multicolumn{2} {|c|} {Single-turn} & \multicolumn {2}{|c|} {Multi-turn} & \multicolumn {2}{|c|} {Multi-turn new application}\\
\hline

Model\textbackslash Metric & $Correlation$ & $F-dissatisfactory$ & $Correlation$ & $F-dissatisfactory$ & $Correlation$ & $F-dissatisfactory$\\
\hline
Lasso & 0.70 $\pm$ 0.01 & 0.69 $\pm$ 0.02 & 0.68 $\pm$ 0.05 & 0.74 $\pm$ 0.06 & 0.62 $\pm$ 0.07 & 0.74 $\pm$ 0.06  \\
\hline
Decision Tree & 0.73 $\pm$ 0.01 & 0.70 $\pm$ 0.03 & 0.67 $\pm$ 0.05 & 0.71 $\pm$ 0.06 & 0.58 $\pm$ 0.07 & 0.73 $\pm$  0.05 \\
\hline
Random Forest &  0.77 $\pm$ 0.01 & 0.74 $\pm$ 0.02 & 0.74 $\pm$ 0.05 & 0.75 $\pm$ 0.05 & 0.61 $\pm$ 0.07 & 0.76 $\pm$  0.05 \\
\hline
G.Boost & \textbf{0.80 $\pm$ 0.01} & \textbf{0.77 $\pm$ 0.02} & \textbf{0.79 $\pm$ 0.04} & \textbf{0.78 $\pm$ 0.05} & \textbf{0.67 $\pm$ 0.06} & \textbf{0.79 $\pm$  0.05} \\
\hline
SVR & 0.73 $\pm$ 0.01 & 0.72 $\pm$ 0.02 & 0.75 $\pm$ 0.04 & 0.71 $\pm$ 0.06 & 0.64 $\pm$ 0.07 & 0.76 $\pm$  0.05 \\
\hline
MLP & 0.75 $\pm$ 0.01 & 0.73 $\pm$ 0.02 & 0.74 $\pm$ 0.04 & 0.73 $\pm$ 0.05 & \textbf{0.67 $\pm$ 0.06} & \textbf{0.79 $\pm$ 0.05} \\
\hline
\end{tabular} 
\egroup
\end{adjustbox}
}
\vspace{-0.0cm}
\caption{Six machine learning regression models performance on turns from single-turn, multi-turn conversations and new multi turn test application. Each cell shows the mean and 95\% bootstrap confidence interval with the highest mean in bold.}
\vspace{-1.0mm}
  \label{tab:single-turn-results}
\end{table*}
Amongst the six models we experimented with, Gradient Boosting Regression achieves superior performance on single-turn domains and multi-turn domains ($r$ = {\scriptsize $\sim$}$0.79$ and F-dissatisfaction = $0.77$). On the new multi-turn application, both Gradient Boosting Regression and MLP models achieve better performance ($r$ = $0.67$ and F-dissatisfaction = $0.79$) in comparison to the other four models we experimented with (results in Table \ref{tab:single-turn-results}).

\begin{table*}[h!]
\captionsetup{font=small}
\centering
\resizebox{0.60\textheight}{!}{
  \begin{adjustbox}{max width=\textwidth}
  \bgroup
  \def\arraystretch{1.1}
\begin{tabular}{|c|c|c|c|c|c|c|}
\hline
\multicolumn {1}{|c}{} & \multicolumn{2} {|c|} {Single-turn} & \multicolumn {2}{|c|} {Multi-turn} & \multicolumn {2}{|c|} {Multi-turn new application}\\
\hline

Features Removed\textbackslash Metric & $Correlation$ & $F-dissatisfactory$ & $Correlation$ & $F-dissatisfactory$ & $Correlation$ & $F-dissatisfactory$\\
\hline
None & 0.80 $\pm$ 0.01 & 0.77 $\pm$ 0.02 & 0.79 $\pm$ 0.04 & 0.78 $\pm$ 0.05 & 0.67 $\pm$ 0.06 & 0.79 $\pm$ 0.05 \\
\hline
Aggegate topic Popularity& \textbf{0.74 $\pm$ 0.02} & \textbf{0.72 $\pm$ 0.02} & 0.75 $\pm$ 0.04 & 0.72 $\pm$ 0.06 & 0.62 $\pm$ 0.07 & 0.74 $\pm$ 0.06  \\
\hline
Un-actionable Request identifier & \textbf{0.76 $\pm$ 0.01} & 0.74 $\pm$ 0.02 & 0.78 $\pm$ 0.05 & 0.77 $\pm$ 0.06 & \textbf{0.50 $\pm$ 0.08} & 0.70 $\pm$  0.05 \\
\hline
Cohesion between request and response &  0.78 $\pm$ 0.01 & 0.77 $\pm$ 0.02 & 0.76 $\pm$ 0.04 & 0.77 $\pm$ 0.05 & 0.63 $\pm$ 0.07 & 0.77 $\pm$  0.06 \\
\hline
Topic diversity & 0.79 $\pm$ 0.01 & 0.77 $\pm$ 0.01 & 0.77 $\pm$ 0.04 & 0.75 $\pm$ 0.05 &  0.67 $\pm$ 0.05 & 0.78 $\pm$  0.05 \\
\hline
User request paraphrasing& 0.78 $\pm$ 0.01 & 0.76 $\pm$ 0.02 & 0.78 $\pm$ 0.04 & 0.76 $\pm$ 0.05 & 0.65 $\pm$ 0.05 & 0.76 $\pm$  0.05 \\
\hline
\end{tabular} 
\egroup
\end{adjustbox}
}
\vspace{-0.0cm}
\caption{Ablation study results on turn-level satisfaction estimation performance. Each cell shows the mean and 95\% bootstrap confidence interval with the highest mean in bold. Performance is measured using Linear correlation between predicted and annotated RQ ratings and F-score on dissatisfactory class (F-dissatisfactory). Bold highlights indicate statistically significant difference in performance.}
\vspace{-1.0mm}
  \label{tab:ablation-study-results}
\end{table*}

Based on ablation study using Gradient boosting regression model, we found that the new features improved every single metric on the test set. On single-turn dialogues, features corresponding to ``aggregate topic popularity'' caused largest statistically significant {\scriptsize $\sim$}$7\%$ relative improvement in linear correlation (0.741 $\rightarrow$ 0.796) and F-dissatisfaction (0.72 $\rightarrow$ 0.77) scores. On the new multi-turn application, {\scriptsize $\sim$}$35\%$ relative improvement in linear correlation (0.496 $\rightarrow$ 0.67) shows significant impact of ``Un-actionable request'' feature on generalization performance (Table\ref{tab:ablation-study-results}). The five new feature sets we introduced occur in the top 10 sets of important features (Appendix Table \ref{top-feature-weights}) returned by model based on their computed importance score.

\subsection{Dialogue-level satisfaction estimation results}
\begin{table*}[h!]
\captionsetup{font=small}
\centering
\resizebox{0.60\textheight}{!}{
  \begin{adjustbox}{max width=\textwidth}
  \bgroup
  \def\arraystretch{1.1}
\begin{tabular}{|c|c|c|c|c|c|c|}
\hline
\multicolumn {1}{|c}{} & \multicolumn{3} {|c|} {Without Average-estimated turn-level rating feature} & \multicolumn 3{|c|} {With Average-estimated turn-level rating feature}\\
\hline
Model\textbackslash Metric & $Accuracy$ & $F-dissatisfactory$ & $Correlation$ & $Accuracy$ & $F-dissatisfactory$ & $Correlation$\\
\hline
Lasso & \textbf{0.73 $\pm$ 0.05} & \textbf{0.64 $\pm$ 0.09} & 0.45 $\pm$ 0.13 & 0.78 $\pm$ 0.07 & 0.71 $\pm$ 0.09 & 0.58 $\pm$ 0.11  \\
\hline
Decision Tree & 0.64 $\pm$ 0.11 & 0.56 $\pm$ 0.12 & 0.31 $\pm$ 0.11 & 0.73 $\pm$ 0.06 & 0.63 $\pm$ 0.08 & 0.52 $\pm$  0.10 \\
\hline
Random Forest &  0.67 $\pm$ 0.06 & 0.61 $\pm$ 0.12 & 0.47 $\pm$ 0.11 & 0.71 $\pm$ 0.07 & 0.64 $\pm$ 0.11 & 0.58 $\pm$  0.10 \\
\hline
G.Boost & 0.68 $\pm$ 0.05 & 0.61 $\pm$ 0.12 & \textbf{0.58 $\pm$ 0.11} & \textbf{0.79 $\pm$ 0.07} & \textbf{0.73 $\pm$ 0.08} & \textbf{0.60 $\pm$  0.09} \\
\hline
\end{tabular} 
\egroup
\end{adjustbox}
}
\vspace{-0.0cm}
\caption{Dialogue level satisfaction estimation performance obtained by four machine learning models with and without Average estimated turn-level satisfaction rating as feature. Each cell shows the mean and 95\% bootstrap confidence interval with the highest mean in bold. Wider confidence intervals are due to small sample size.}
\vspace{-1.0mm}
  \label{tab:dialogue-level-results}
\end{table*}
Across all four models we experimented with, we observed an improvement in dialogue level satisfaction estimation results by including average estimated turn-level satisfaction rating as a feature (results in Table \ref{tab:dialogue-level-results}). Turn-level satisfaction ratings were estimated using the Gradient Boosting Regression model which is described in the previous section (Section \ref{sec:turn-level-satisfaction-results}). Best performing dialogue-level Gradient Boosting Regression model achieved $16\%$  relative improvement (68\%$\rightarrow$79\%) in binary ``satisfactory/dissatisfactory'' class prediction accuracy and {\scriptsize $\sim$}$3.5\%$ relative improvement (0.58$\rightarrow$0.60) in correlation between predicted and true dialogue-level labels. Based on feature importance scores returned by the Gradient Boosting regression model, average predicted turn-level satisfaction rating score is the most important feature for predicting dialogue-level satisfaction rating, followed by average ASR scores and average aggregate domain popularity (top 10 features in Appendix Table \ref{top-feature-weights-dialogue}).


\section{Conclusion}
In this paper, we described a user-centric and domain-independent approach for evaluating user satisfaction in multi-domain conversations users have with an AI assistant. We introduced Response Quality (RQ) annotation scheme which is highly correlated ($r$ = $0.76$) with explicit turn level user satisfaction ratings. By designing five additional new features, we achieved a high linear correlation of {\scriptsize $\sim$}0.79 between annotated RQ and predicted User Satisfaction ratings with Gradient Boosting Regression as the User Satisfaction Estimation (USE) model, for both single-turn and multi-turn dialogues. Gradient Boosting Regression and Multi Layer Perceptron (MLP) models generalized to new domain better ($r$ = $0.67$) than other models. By including turn-level satisfaction prediction as features, we observed a relative $16\%$ improvement (68\% $\rightarrow$ 79\%) improvement in binary ``dissatisfactory/satisfactory'' class prediction accuracy of a domain-independent dialogue-level satisfaction estimation model. The Gradient Boost Regression based dialogue-level model estimated explicit dialogue-level ratings provided by novice and experienced users on conversations spanning $14$ domains and achieved significant (at 95\% confidence interval) $0.60$ correlation between predicted and true user ratings.
 
With statistically significant {\scriptsize $\sim$}$7\%$ and {\scriptsize $\sim$}$35\%$ relative improvement in linear correlation on existing domains and new multi-turn application respectively, our ablation study supported our hypothesis that the new features improve model prediction performance. On multi-domain conversations, we plan to explore the use of Deep Neural Net models to reduce handcrafting of features \citep*{Rach2017InteractionQE} and to jointly estimate end user satisfaction at turn and dialogue level, though these models reduce interpretability. To learn dialogue policies using reinforcement learning, we plan to experiment with proposed RQ based User Satisfaction metric as an alternative for reward modeling.

\subsubsection*{Acknowledgments}
We thank Alborz Geramifard and NeurIPS Conversation AI workshop reviewers for their insightful feedback and guidance. We also thank Longshaokan Wang, Swanand Joshi and Joshua Levy for helping with data procurement and turn level satisfaction estimation setup. Finally, we thank Alison Bauter Engel, Kate Ridgeway and Alexa Data Services-RAMP team for their significant help with user studies and data annotations.



\newpage
\small
\bibliographystyle{named}
\bibliography{neurips_2019.bib}

\begin{thebibliography}{}

\bibitem[\protect\citeauthoryear{Agrawal \bgroup \em et al.\egroup
  }{1993}]{Agrawal:1993:MAR:170036.170072}
Rakesh Agrawal, Tomasz Imieli\'{n}ski, and Arun Swami.
\newblock Mining association rules between sets of items in large databases.
\newblock {\em SIGMOD Rec.}, 22(2):207--216, June 1993.

\bibitem[\protect\citeauthoryear{Asri \bgroup \em et al.\egroup
  }{2014}]{6854195}
L.~E. Asri, H.~Khouzaimi, R.~Laroche, and O.~Pietquin.
\newblock Ordinal regression for interaction quality prediction.
\newblock In {\em 2014 IEEE International Conference on Acoustics, Speech and
  Signal Processing (ICASSP)}, pages 3221--3225, May 2014.

\bibitem[\protect\citeauthoryear{Breiman}{2001}]{breiman2001random}
Leo Breiman.
\newblock Random forests.
\newblock {\em Machine learning}, 45(1):5--32, 2001.

\bibitem[\protect\citeauthoryear{Deriu \bgroup \em et al.\egroup
  }{2019}]{DBLP:journals/corr/abs-1905-04071}
Jan Deriu, {\'{A}}lvaro Rodrigo, Arantxa Otegi, Guillermo Echegoyen, Sophie
  Rosset, Eneko Agirre, and Mark Cieliebak.
\newblock Survey on evaluation methods for dialogue systems.
\newblock {\em CoRR}, abs/1905.04071, 2019.

\bibitem[\protect\citeauthoryear{Drucker \bgroup \em et al.\egroup
  }{1997}]{NIPS1996_1238}
Harris Drucker, Christopher J.~C. Burges, Linda Kaufman, Alex~J. Smola, and
  Vladimir Vapnik.
\newblock Support vector regression machines.
\newblock In M.~C. Mozer, M.~I. Jordan, and T.~Petsche, editors, {\em Advances
  in Neural Information Processing Systems 9}, pages 155--161. MIT Press, 1997.

\bibitem[\protect\citeauthoryear{Engelbrecht \bgroup \em et al.\egroup
  }{2009}]{engelbrecht-etal-2009-modeling}
Klaus-Peter Engelbrecht, Florian G{\"o}dde, Felix Hartard, Hamed Ketabdar, and
  Sebastian M{\"o}ller.
\newblock Modeling user satisfaction with hidden {M}arkov models.
\newblock In {\em Proceedings of the {SIGDIAL} 2009 Conference}, pages
  170--177, London, UK, September 2009. Association for Computational
  Linguistics.

\bibitem[\protect\citeauthoryear{Friedman}{2001}]{friedman2001greedy}
Jerome~H Friedman.
\newblock Greedy function approximation: a gradient boosting machine.
\newblock {\em Annals of statistics}, pages 1189--1232, 2001.

\bibitem[\protect\citeauthoryear{Gardner and
  Dorling}{1998}]{gardner1998artificial}
MW~Gardner and SR~Dorling.
\newblock {Artificial neural networks (the multilayer perceptron)---A review of
  applications in the atmospheric sciences}.
\newblock {\em Atmospheric Environment}, 32(14-15):2627--2636, 1998.

\bibitem[\protect\citeauthoryear{Higashinaka \bgroup \em et al.\egroup
  }{2010}]{Higashinaka2010IssuesIP}
Ryuichiro Higashinaka, Yasuhiro Minami, Kohji Dohsaka, and Toyomi Meguro.
\newblock Issues in predicting user satisfaction transitions in dialogues:
  Individual differences, evaluation criteria, and prediction models.
\newblock In {\em IWSDS}, 2010.

\bibitem[\protect\citeauthoryear{Pragst \bgroup \em et al.\egroup
  }{2017}]{Pragst2017}
Louisa Pragst, Stefan Ultes, and Wolfgang Minker.
\newblock {\em Recurrent Neural Network Interaction Quality Estimation}, pages
  381--393.
\newblock Springer Singapore, Singapore, 2017.

\bibitem[\protect\citeauthoryear{Rach \bgroup \em et al.\egroup
  }{2017}]{Rach2017InteractionQE}
Niklas Rach, Wolfgang Minker, and Stefan Ultes.
\newblock Interaction quality estimation using long short-term memories.
\newblock In {\em SIGDIAL Conference}, 2017.

\bibitem[\protect\citeauthoryear{Schmitt \bgroup \em et al.\egroup
  }{2011}]{Schmitt2011ModelingAP}
Alexander Schmitt, Benjamin Schatz, and Wolfgang Minker.
\newblock Modeling and predicting quality in spoken human-computer interaction.
\newblock In {\em SIGDIAL Conference}, 2011.

\bibitem[\protect\citeauthoryear{Schmitt \bgroup \em et al.\egroup
  }{2012}]{SCHMITT12.333}
Alexander Schmitt, Stefan Ultes, and Wolfgang Minker.
\newblock A parameterized and annotated spoken dialog corpus of the cmu let's
  go bus information system.
\newblock In Nicoletta Calzolari~(Conference Chair), Khalid Choukri, Thierry
  Declerck, Mehmet~U?ur Do?an, Bente Maegaard, Joseph Mariani, Asuncion Moreno,
  Jan Odijk, and Stelios Piperidis, editors, {\em Proceedings of the Eight
  International Conference on Language Resources and Evaluation (LREC'12)},
  Istanbul, Turkey, may 2012. European Language Resources Association (ELRA).

\bibitem[\protect\citeauthoryear{Tibshirani}{1996}]{10.2307/2346178}
Robert Tibshirani.
\newblock Regression shrinkage and selection via the lasso.
\newblock {\em Journal of the Royal Statistical Society. Series B
  (Methodological)}, 58(1):267--288, 1996.

\bibitem[\protect\citeauthoryear{Ultes \bgroup \em et al.\egroup
  }{2014}]{10.1007/978-1-4614-8280-2_27}
Stefan Ultes, Robert ElChab, and Wolfgang Minker.
\newblock Application and evaluation of a conditioned hidden markov model for
  estimating interaction quality of spoken dialogue systems.
\newblock In Joseph Mariani, Sophie Rosset, Martine Garnier-Rizet, and Laurence
  Devillers, editors, {\em Natural Interaction with Robots, Knowbots and
  Smartphones}, pages 303--312, New York, NY, 2014. Springer New York.

\bibitem[\protect\citeauthoryear{Ultes \bgroup \em et al.\egroup
  }{2017}]{Ultes2017DomainIndependentUS}
Stefan Ultes, Pawel Budzianowski, I{\~n}igo Casanueva, Nikola Mrksic,
  Lina~Maria Rojas-Barahona, Pei hao Su, Tsung-Hsien Wen, Milica Gasic, and
  Steve~J. Young.
\newblock Domain-independent user satisfaction reward estimation for dialogue
  policy learning.
\newblock In {\em INTERSPEECH}, 2017.

\bibitem[\protect\citeauthoryear{Walker \bgroup \em et al.\egroup
  }{2000}]{Walker:2000:TDG:973935.973945}
Marilyn Walker, Candace Kamm, and Diane Litman.
\newblock Towards developing general models of usability with paradise.
\newblock {\em Nat. Lang. Eng.}, 6(3-4):363--377, September 2000.

\end{thebibliography}

\newpage
\appendix

\onecolumn
\section{Appendices}
\label{sec:appendix}
\begin{table*}[h!]
\captionsetup{font=small}
\small
  \centering
  \begin{adjustbox}{max width=\textwidth}
  \bgroup
  \def\arraystretch{1.0}
  \begin{tabularx}{\linewidth}{lX}
    \toprule
    \multicolumn{1}{c}{\textbf{Domain/Application}} & \multicolumn{1}{c}{\textbf{Goal}} \\
    \midrule
    Multi-turn conversational bot for movies & Get ratings of two other movies directed by the director of a movie you liked watching recently. \\
    Multi-turn conversational bot for movies & Ask for movie recommendations and navigate through them until you got a recommendation you liked.\\
    Multi-turn conversational bot for movies $\rightarrow$ Music & Find details about the cast of your favorite movie and play the movie's soundtrack.\\
    Ticket-booking third party application & Get a list of theaters near you and find out what is playing in at least 2 theaters returned by the application for the same query. \\
    Music$\rightarrow$Knowledge & Find song by artist and find out in which movie it was used\\
    Music$\rightarrow$Music & Find a popular album by an artist and play it\\
    Knowledge$\rightarrow$Local Search & Ask for a fact about a historical monument and find its address\\
    Recipe & Ask for a general type of recipe (e.g. cookies,  pasta dishes, etc.), choose a recipe from the list using your voice, and then add the ingredients to your shopping list. \\
    Ticket-booking third party application & Try purchasing ticket for a movie playing in theaters near you. \\
    Ticket-booking third party application & Find what's playing in a theater outside your current city (or outside the default city from your device)\\
    Notification & Add an appointment to the calendar and then change the time. (requires the user to first cancel and then create a new appointment)\\
    \bottomrule   
  \end{tabularx}
  \egroup
  \end{adjustbox}
  \caption{Example Multi-turn goals for user study}
  \label{user-study-goals}
\end{table*}

\begin{table*}[h!]
\captionsetup{font=small}
\centering
    \resizebox{0.60\textheight}{!}{
      \bgroup
  \begin{adjustbox}{max width=\textwidth}
  \def\arraystretch{1.1}
  \begin{tabularx}{\textwidth}{X X}
    \toprule
    \multicolumn{1}{c}{\textbf{Model}}  & \multicolumn{1}{c}{\textbf{Hyper parameter - Optimal value}}  \\
    \midrule
   Lasso & alpha:\,$0.001$\\
   Decision Trees &  max-depth:\,$33$,\,min-samples-leaf:\,$31$,\,min-samples-split:\,$23$ \\
   Random Forest &  max-depth:\,$49$,\,min-samples-leaf:\,$11$,\,min-samples-split:\,$27$ \\
   Gradient Boosting Decision Trees &  max-depth:\,$23$,\,min-samples-leaf:\,$17$,\,min-samples-split:\,$59$ \\
   SVR & c:\,$2$,\,gamma:\,$0.024$\\
   MLP & n-layers:\,$3$,\,batch-size:\,$128$,\,hidden size\,:\,$100$,\,solver:\,`sgd'\,activation:\,`relu'\\  
  \bottomrule
  \end{tabularx}
  \end{adjustbox}
        \egroup
  }
   \caption{Optimal Hyper parameter values used for training turn-level User Satisfaction Estimation (USE) models.}
  \label{optimal-hyper-parameter}
\end{table*}

\pagebreak
\begin{table*}[h!]
\captionsetup{font=small}
  \centering
   \resizebox{0.60\textheight}{!}{
  \begin{adjustbox}{max width=\textwidth}
  \bgroup
  \def\arraystretch{1.1}
  \begin{tabularx}{\linewidth}{XX}
    \toprule
    \multicolumn{1}{c}{\textbf{Feature set description}} & \multicolumn{1}{c}{\textbf{Turn(s) the feature set is computed on}} \\
    \midrule
ASR \& NLU Confidence scores & \hspace{10mm} $t_{n}^u$ \\
Length of user request and system Response & \hspace{10mm} $t_{n}^u$, $t_{n}^s$ \\ 
Time between consecutive user requests & \hspace{10mm} $t_{n}^u$-$t_{n+1}^u$  \\
\textbf{Aggregate - Intent Popularity}  & \hspace{10mm} $t_{n}^u$ \\
\textbf{Un-actionable user request} & \hspace{10mm} $t_{n}^s$ \\
\textbf{Cohesion between system response and user request}  & \hspace{10mm} $t_{n}^u$, $t_{n}^s$ \\
 Length of dialogue  & \hspace{10mm} $t_{0}$-$t_{n}$ \\
\textbf{Topic diversity} & \hspace{10mm} $t_{0}^u$-$t_{n}^u$\\
\textbf{User paraphrasing his/her request} & \hspace{10mm} $t_{n}^u$-$t_{n+1}^u$\\
    \bottomrule
  \end{tabularx}
  \egroup
  \end{adjustbox}
  }
   \caption{Based on turn-level Gradient Boosting Regression model's output, top 10 feature sets \protect\footnotemark ranked by Importance score. In bold are the new feature sets we introduced.}
   \vspace{-2.5mm}
  \label{top-feature-weights}
\end{table*}
\footnotetext{For confidentiality we are not mentioning the entire list of feature sets used in the model.}

\begin{table}[h!]
\captionsetup{font=small}
\centering
    \resizebox{0.60\textheight}{!}{
      \bgroup
  \begin{adjustbox}{max width=\textwidth}
  \def\arraystretch{1.1}
  \begin{tabularx}{\textwidth}{X X}
    \toprule
    \multicolumn{1}{c}{\textbf{Model}}  & \multicolumn{1}{c}{\textbf{Hyper parameter - Optimal value}}  \\
    \midrule
   Lasso & alpha:\,$0.01$\\
   Decision Trees &  max-depth:\,$2$,\,min-samples-leaf:\,$5$,\,min-samples-split:\,$2$ \\
   Random Forest &  max-depth:\,$4$,\,min-samples-leaf:\,$8$,\,min-samples-split:\,$13$ \\
   Gradient Boosting Decision Trees &  max-depth:\,$2$,\,min-samples-leaf:\,$8$,\,min-samples-split:\,$17$ \\
  \bottomrule
  \end{tabularx}
  \end{adjustbox}
    \egroup
  }
\caption{Optimal Hyper parameter values used for training dialogue-level satisfaction Estimation models.}
\label{optimal-hyper-parameter-dialogue}
\end{table}

\begin{table}[h!]
\captionsetup{font=small}
  \centering
   \resizebox{0.60\textheight}{!}{
  \begin{adjustbox}{max width=\textwidth}
  \bgroup
  \def\arraystretch{1.1}
  \begin{tabularx}{\linewidth}{XX}
    \toprule
    \multicolumn{1}{c}{\textbf{Feature set description}} & \multicolumn{1}{c}{\textbf{Turn(s) the feature set is computed on}} \\
    \midrule
Average estimated turn-level satisfaction ratings & \hspace{10mm} $t_{0}$-$t_{N}$ \\
Average ASR confidence & \hspace{10mm} $t_{0}$-$t_{N}$ \\ 
Last turn's ASR confidence & \hspace{10mm} $t_{N}$ \\
Average aggregate - domain popularity & \hspace{10mm} $t_{0}$-$t_{N}$ \\
Number of question prompts from the system side & \hspace{10mm} $t_{0}^s$-$t_{N}^s$  \\
Average time difference between consecutive utterances & \hspace{10mm} $t_{0}^u$-$t_{N}^u$  \\
Average NLU confidence & \hspace{10mm} $t_{0}$-$t_{N}$ \\ 
Length of the dialogue & \hspace{10mm} $t_{0}$-$t_{N}$ \\ 
Last turn's aggregate intent popularity & \hspace{10mm} $t_{N}$ \\ 
Average aggregate - intent popularity & \hspace{10mm} $t_{0}$-$t_{N}$ \\
   \bottomrule
  \end{tabularx}
  \egroup
  \end{adjustbox}
  }
   \caption{Based on dialogue-level Gradient Boosting Regression model output, top 10 features, \protect\footnotemark ranked by Importance score.}   \vspace{-2.5mm}
  \label{top-feature-weights-dialogue}
\end{table}
\footnotetext{For confidentiality we are not mentioning the entire list of features used in the model.}

\label{sec:appendix}
\end{document}